\documentclass[journal,twoside,web]{ieeecolor2}
\usepackage{generic}
\usepackage{cite}
\usepackage{amsmath,amssymb,amsfonts}
\usepackage{algorithm}
\usepackage{algorithmic}
\usepackage{graphicx}
\usepackage{textcomp}
\usepackage{gensymb}
\usepackage{booktabs}
\usepackage{array}
\usepackage{xcolor}

\def\BibTeX{{\rm B\kern-.05em{\sc i\kern-.025em b}\kern-.08em
    T\kern-.1667em\lower.7ex\hbox{E}\kern-.125emX}}
\begin{document}
\title{\textbf{RAUL: Reference-Assisted Ureteroscopy Localization for Skill Assessment}}
\author{Fangjie Li, Mai Bui, Charan Mohan, Michael Miga, Matthieu Chabanas, Nicholas Kavoussi, and \\Jie Ying Wu 
\thanks{This paper was submitted on August 17, 2026. This study was supported in part by the NIBIB of the NIH Grant 1R21EB035783.}
\thanks{Fangjie Li is with the Department of Biomedical Engineering, Vanderbilt University, Nashville, TN 37240 USA (e-mail: fangjie.li@vanderbilt.edu)}
\thanks{Mai Bui is with the Department of Computer Science, Vanderbilt University, Nashville, TN 37240 USA (e-mail: mai.n.bui@vanderbilt.edu)}
\thanks{Charan Mohan is with the Department of Urology, Vanderbilt University Medical Center, Nashville, TN 37232 USA (e-mail: charan.mohan@vumc.org)}
\thanks{Michael Miga is with the Department of Biomedical Engineering, Vanderbilt University, Nashville, TN 37240 USA (e-mail: michael.i.miga@vanderbilt.edu)}
\thanks{Matthieu Chabanas is with the Department of Computer Science, Vanderbilt University, Nashville, TN 37240 USA (e-mail: matthieu.chabanas@vanderbilt.edu)}
\thanks{Nicholas Kavoussi is with the Department of Urology, Vanderbilt University Medical Center, Nashville, TN 37232 USA (e-mail: nicholas.l.kavoussi@vumc.org)}
\thanks{Jie Ying Wu is with the Department of Computer Science, Vanderbilt University, Nashville, TN 37240 USA (e-mail: JieYing.Wu@vanderbilt.edu)}
}

\maketitle

\begin{abstract}
Objective: Incomplete navigation of anatomy during ureteroscopic kidney stone surgeries can contribute to repeat interventions. While skilled surgeons have lower reintervention rates, there are no objective metrics to quantify scope-navigation performance to evaluate when a trainee becomes skilled. This work aims to recover ureteroscope trajectories from endoscopic video and derive navigation metrics to quantify differences in skill.

Methods: We propose RAUL, a reference-assisted reconstruction framework for recovering ureteroscope trajectories from ureteroscope videos only in phantoms. For each phantom, we use a slow, high-quality reference exploration video to generate a reference reconstruction. We localize subsequent exploration videos against this reference. We evaluate localization accuracy against electromagnetically tracked scope pose. We compute navigation metrics from phantom exploration trajectories to compare surgical residents across experience levels.

Results: The proposed reference-assisted framework achieves a mean translation root mean square error of $0.5 \pm 0.1$ mm across 9 phantoms. Compared to standard Structure-from-Motion (SfM),  the proposed pipeline increases frame-wise localization coverage from $50.5 \pm 14.9\%$ to $86.1 \pm 7.2\%$ of all video frames. The reconstructed trajectories revealed significant differences between high- and low-experience trainees in established navigation metrics.

Conclusion: RAUL enables substantially more complete recovery of ureteroscope trajectories from videos compared to standard SfM pipelines, enabling trajectory-based skill assessment without additional tracking equipment.

Significance: To the best of our knowledge, this is the first use of video-only recovery of ureteroscope trajectories without external tracking sensors for skill assessment, supporting scalable automated assessment of ureteroscopy navigation skill.
\end{abstract}


\section{Introduction}
\label{sec:introduction}
Among the over 100,000 patients undergoing flexible ureteroscopy (fURS) kidney stone surgeries annually in the U.S., approximately a quarter require repeat stone surgeries within 20 months~\cite{Brain2021}. This is largely due to residual fragments not cleared, which can cause pain, injury, and infectious complications~\cite{Khanna2020}. Complete stone clearance requires thorough visualization and navigation of the renal collecting system with a ureteroscope inserted through the ureter. This can be challenging given the complex anatomy and the limited field of view of the ureteroscope. These are challenging skills, closely linked to surgeon experience. Expert stone surgeons achieved substantially higher stone-free rates than less experienced surgeons~\cite{Wolff2019}. 


Despite the importance of surgeon experience, developing proficiency in fURS remains challenging. Prior work suggests that approximately 50 cases may be required for resident performance to mature~\cite{csahin2023retrograde}. Residents typically learn how to perform fURS in live surgeries, which limits opportunities to receive feedback due to patient-safety and operative-time constraints. Thus, feedback is often qualitative, delayed, and difficult to standardize~\cite{Vedula2017}. Feedback limitations motivate the development of automated and objective systems for skill assessment and training outside of the operating room (OR). In particular, objective measures of scope navigation performance could enable trainees to identify inefficient exploration patterns and track their progression toward expert-level performance~\cite{ma2022tailored}.

Such automated assessment systems and metrics exist for robotic surgeries. However, they usually require tool pose data~\cite{wang2018deep, Fawaz2018KinematicCNN}, which fURS lacks. Ureteroscopy simulators also exist~\cite{AcarAyberk2025NNAR,Brunckhorst2015,Aydin2016}, but those that support automated assessment~\cite{Brunckhorst2015,Aydin2016} require costly phantoms with additional sensors and hardware, limiting their scalability and widespread adoption. 

These limitations motivate a scalable, cost-effective approach for retrieving ureteroscope trajectory from ureteroscope video inherently generated during fURS. Such trajectories would enable established tool-motion-based skill assessment methods to be applied to fURS without specialized tracking hardware. 

The advancement of computer vision-based methods such as Simultaneous Localization and Mapping (SLAM), Structure from Motion (SfM), and deep learning-based approaches shows promise in video-based ureteroscopic trajectory estimation~\cite{OlivaMaza2023ORBSLAMUreteroscopy,schmidt2024tracking,acar2026perseus,acar2025monocular}. However, they can have limited robustness on poor-quality surgical videos~\cite{widya20193d}, such as those shown in Fig.~\ref{fig:sfm_failure_examples}. Building on vision-based 3D reconstruction, we propose RAUL, a video-only reference-assisted framework for quantitative analysis of ureteroscope navigation behavior. Given ureteroscope videos, the proposed framework estimates ureteroscope poses and reconstructs the explored anatomy by matching the video against a prior reference reconstruction from expert navigation. This provides dense pose reconstruction that, when combined with a low-cost phantom~\cite{AcarAyberk2025NNAR}, can enable the large-scale automated assessment outside of the OR with limited reliance on additional hardware, reducing the economic and adoption burden. 

Using the reconstructed trajectories, we compute commonly used navigation metrics capturing ureteroscope motion smoothness and navigation efficiency. These metrics capture aspects of technical performance related to fluid instrument handling, economy of motion, and efficient task execution, which are qualitatively assessed in the established Objective Structured Assessment of Technical Skill (OSATS)~\cite{Martin1997OSATS}. We evaluate the framework on a phantom exploration dataset by testing whether the metrics differ across groups with different levels of clinical experience, thereby assessing whether they capture experience-related differences in ureteroscope navigation.

Concretely, we introduce the following contributions in this paper:
\begin{enumerate}
    \item We introduce RAUL, a reference-assisted reconstruction framework for recovering ureteroscope poses from visually challenging endoscopic videos.
    
    \item We use the ureteroscope poses to compute navigation metrics, including trajectory smoothness, path length, and efficiency.
    
    \item We show a significant difference in navigation metrics between resident groups with different experience levels, across ten users. This suggests that the estimated ureteroscope poses can be used for skill assessment and feedback. 

\end{enumerate}

\section{Related Works}
\subsection{Manual Skill Assessment}
Objective surgical skill assessment has traditionally relied on structured manual evaluation frameworks such as the OSATS, which provides interpretable assessment of technical performance but requires expert review and is difficult to scale~\cite{Martin1997OSATS}. Crowd-Sourced Assessment of Technical Skills (C-SATS) partially addresses scalability by showing that trained crowd raters can evaluate surgical videos~\cite{Holst2015CSATS}. However, both expert- and crowd-based video assessments rely primarily on subjective ratings of perceived performance quality. Moreover, prior work found that manual review of ureteroscopic video has not reliably distinguished surgical expertise, and crowd-sourced scores correlated poorly with expert scores for ureteroscopic video assessment~\cite{Conti2019}. 

\subsection{Robotic Surgery Skill Assessment}
Surgical skill assessment in robot-assisted minimally invasive surgery provides synchronized endoscopic video and tool pose data. The JIGSAWS dataset established a common benchmark for robotic gesture and skill analysis using synchronized video and pose data from standardized dry-lab tasks such as suturing, needle passing, and knot tying~\cite{Gao2014JIGSAWS}. Subsequent methods have used robot-motion time series, handcrafted trajectory features, deep temporal models, and image-based classifiers to infer skill. For example, Wang and Fey trained a convolutional neural network on multivariate robot trajectory~\cite{wang2018deep}, while Fawaz \emph{et al.} used convolutional models with class activation maps to identify task segments influential for skill prediction~\cite{Fawaz2018KinematicCNN}. Video-based approaches have also been explored, including image-based skill classification in robot-assisted minimally invasive surgery~\cite{Lajko2021EndoscopicSkillAssessment} and transformer-based surgical activity and skill decoding from robotic surgery videos~\cite{Kiyasseh2023SAIS, Laughlin}.

These studies show that video and pose signals contain discriminative information about surgical proficiency. Additionally, feedback based on these signals is shown to accelerate robotic surgery skill acquisition~\cite{ma2022tailored}. However, adapting these approaches to fURS presents several challenges. Learned video-based skill classification would require a sufficiently large labeled dataset spanning different experience levels, anatomies, and procedural conditions, which is not currently available for ureteroscopy. The aforementioned unreliability of manual video assessment further limits learning-based skill classification~\cite{Conti2019}. In contrast, trajectory-based assessment computes established and interpretable navigation metrics~\cite{Singh2021MotionSmoothnessCannulation}, allowing assessment without the need for task-specific training. Unfortunately, its use in fURS is currently limited by the absence of routinely available ureteroscope pose data. We address this limitation by recovering trajectories directly from endoscopic video and evaluating navigation over complete explorations of realistic kidney anatomy.

\subsection{Objective Assessment in Non-Robotic Endoscopy and Ureteroscopy}
Prior work has explored external tracking, motion sensors, gaze tracking, and video-based motion analysis to address the lack of tool pose data. Atoum \emph{et al.} tracked and used gaze behavior during ureteroscopic exploration to evaluate task performance~\cite{Atoum}. Singh \emph{et al.} analyzed trajectory quality in a cannulation simulator using external electromagnetic and optical sensing~\cite{Singh2021MotionSmoothnessCannulation}. Most closely related to our work, Valovska \emph{et al.} developed the Composite Ureteroscopy Efficiency Score (CUES), using a simulated ureteroscopic task with motion capture and video-camera-based computer vision algorithms to quantify travel distance, task duration, smoothness, and wall-collision counts~\cite{Valovska2023CUES}.

Together, these studies demonstrate the feasibility of objective skill assessment in non-robotic endoscopy, but rely on auxiliary sensors, external camera views, or instrumentation specific to a particular simulator. These requirements increase system cost and may limit adaptability across training environments and anatomies. Our work instead targets low-cost, video-based ureteroscope localization without additional tracking hardware, supporting objective assessment across anatomically realistic kidney phantoms with varied collecting-system geometries.

\subsection{Endoscopic 3D Reconstruction}
Endoscopic 3D reconstruction and localization have been widely studied for surgical navigation, camera tracking, and scene understanding~\cite{schmidt2024tracking}. SLAM and SfM pipelines have been adapted to endoscopic navigation, including ureteroscopy~\cite{OlivaMaza2023ORBSLAMUreteroscopy}, while recent learning-based methods have advanced endoscopic scene reconstruction and endoscope localization~\cite{Allan2021SCARED,Shao2022AFSfMLearner,guo2025endo3r}.  Nevertheless, reconstruction remains challenging because endoscopic video commonly contains weak texture, specular reflections, rapid motion, narrow fields of view, non-Lambertian appearance, and tissue deformation. Under these conditions, existing methods can lose tracking or reconstruct only subsets of a video, particularly during motion blur, occlusion, poor visibility, or insufficient visual overlap~\cite{widya20193d,azagra2023endomapper}. Additionally, quantitative evaluations commonly use controlled, or short video sequences curated for reliable ground-truth pose or depth estimation~\cite{bobrow2023colonoscopy,Allan2021SCARED,Ozyoruk2021EndoSLAM}. Even datasets containing complete clinical procedures often provide reconstruction benchmarks only for shorter segments with smooth motion and favorable visibility~\cite{azagra2023endomapper}. In contrast, objective trajectory-based skill assessment requires localization across many complete trainee explorations, which often exceed one minute and contain abrupt motion, blur, poor visibility, and repeated views of visually similar anatomy.

Our setting also differs from intraoperative guidance. Skill assessment permits offline processing and repeated exploration of the same anatomy. We can therefore acquire a high-quality reference exploration video and use non-real-time methods to prioritize localization robustness. Together, these differences enable existing reconstruction and localization techniques to be adapted for analyzing many complete exploration videos.

\section{Methods}
\begin{figure*}[!t]
\centerline{\includegraphics[width=0.8\textwidth]{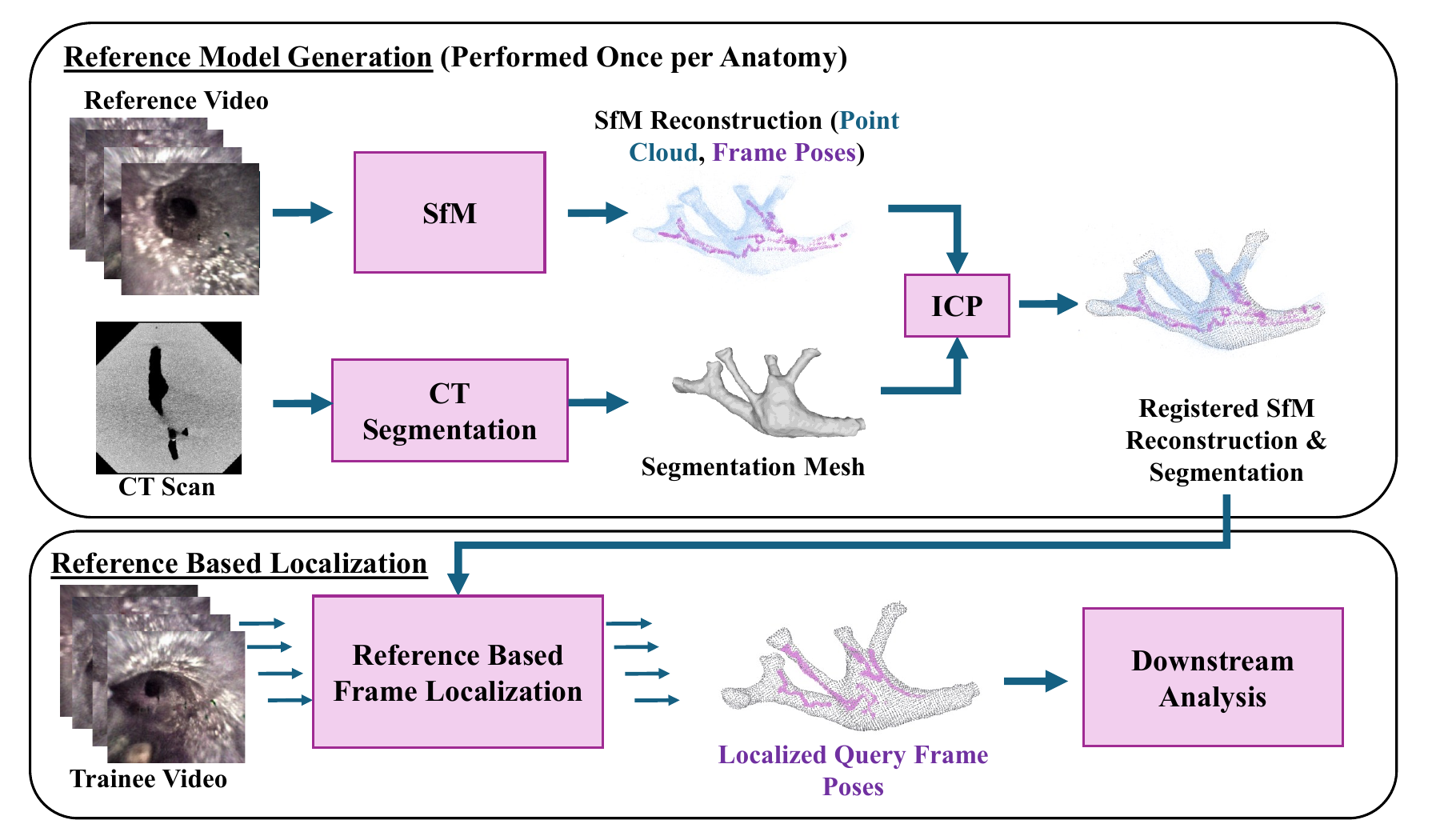}}
\caption{The overall pipeline for combined reconstruction, highlighting the SfM reference reconstruction generation. By using a reference video for each phantom, we are able to improve reconstruction robustness.}
\label{fig:fig1_overall_pipeline}
\end{figure*}

We first introduce RAUL, a reference-assisted ureteroscopy localization workflow designed to maximize trajectory coverage and robustness. For each anatomy, we collect a slow, high-coverage reference exploration video, which is reconstructed into a reusable 3D map. We then localize explorations against this reference map, allowing ureteroscope trajectories from different runs to be recovered in a common anatomical coordinate frame. The overall workflow is outlined in Fig.~\ref{fig:fig1_overall_pipeline}. Second, we use these trajectories to compute established navigation metrics.

\subsection{Robust Reconstruction for Ureteroscope Pose Estimation}

We estimate ureteroscope poses from monocular endoscopic video frames using SfM-based localization. There are three main steps in a standard SfM reconstruction pipeline: 1) covisible video frames are identified, 2) 2D image feature points are extracted from each frame, and corresponding features are matched in covisible images, and 3) the relative ureteroscope pose of each frame and the 3D positions of the features are triangulated and optimized. 

\begin{figure}
\centerline{\includegraphics[width=0.9\columnwidth]{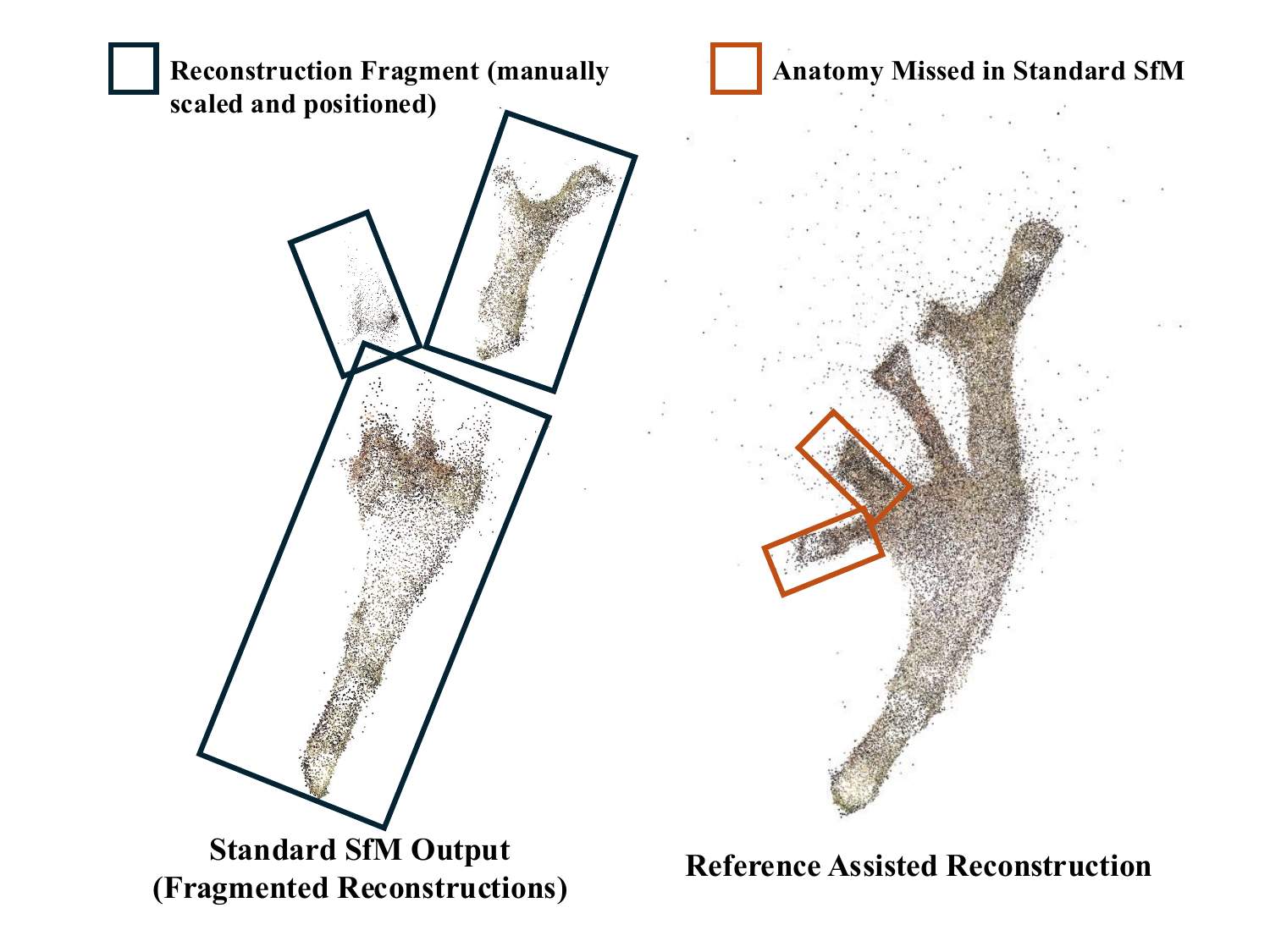}}
\caption{An example standard SfM output case is displayed along with a full reconstruction through the proposed method. Note how the SfM algorithm fails to build a unified map, and misses portions of the full anatomy. Here, the fragmented reconstructions are manually scaled and placed together, a manual process that would further hinder the use of the SfM output for downstream objective skill assessment.}
\label{fig:sfm_output_examples}
\end{figure}

Standalone SfM reconstruction of realistic-speed ureteroscopy exploration videos often produces fragmented or incomplete results. Two common failure modes are shown in Fig.~\ref{fig:sfm_output_examples}: 1) a video may decompose into several local segments as frames within each group have sufficient feature matches to reconstruct a local structure, but matches between segments are insufficient to merge them into a single reconstruction and 2) some groups of frames may fail to reconstruct entirely. These failures make the output difficult to use for downstream trajectory-based skill assessment, because each local reconstruction has an ambiguous scale and pose relative to the others. 

\begin{figure}
    \centering
    \includegraphics[width=0.7\linewidth]{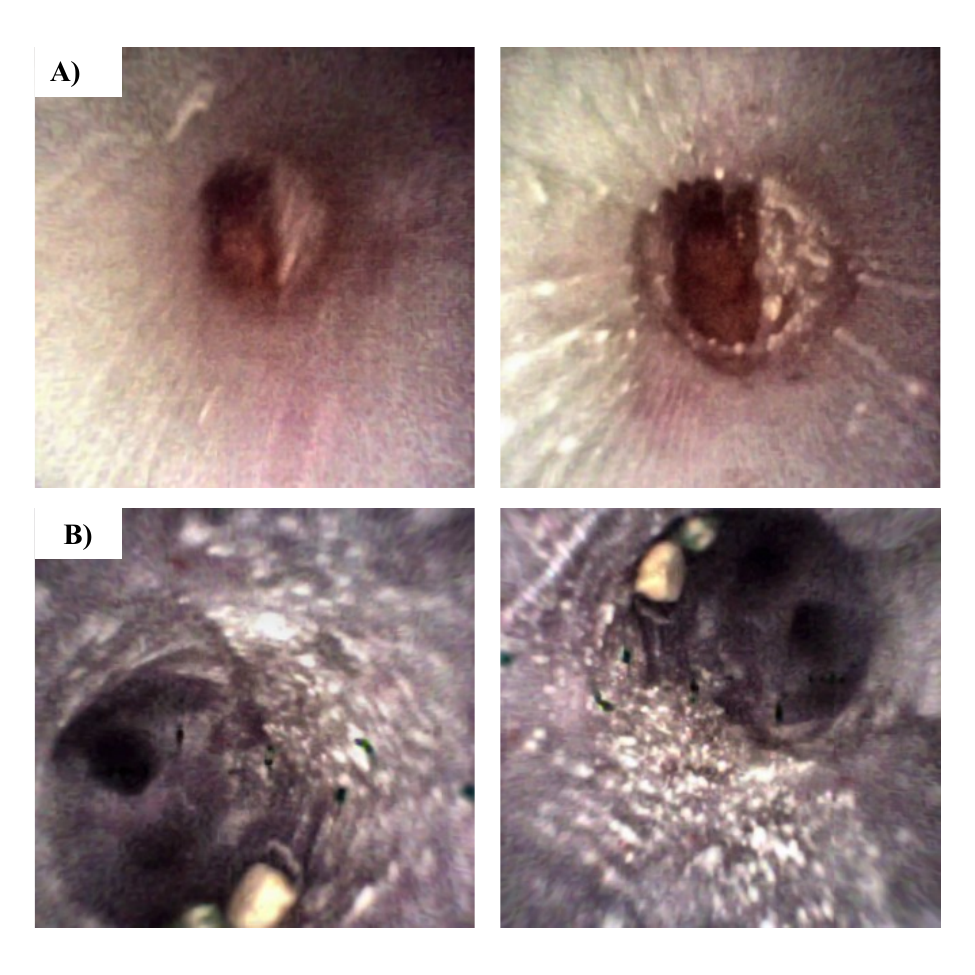}
    \caption{Example images highlighting feature matching failure cases. A) Motion Blur. B) Large In-Plane Rotation. Ureteroscope images have overall low resolution, lack of texture, and specularities.}
    \label{fig:sfm_failure_examples}
\end{figure}

\begin{figure*}[!t]
\centerline{\includegraphics[width=0.8\textwidth]{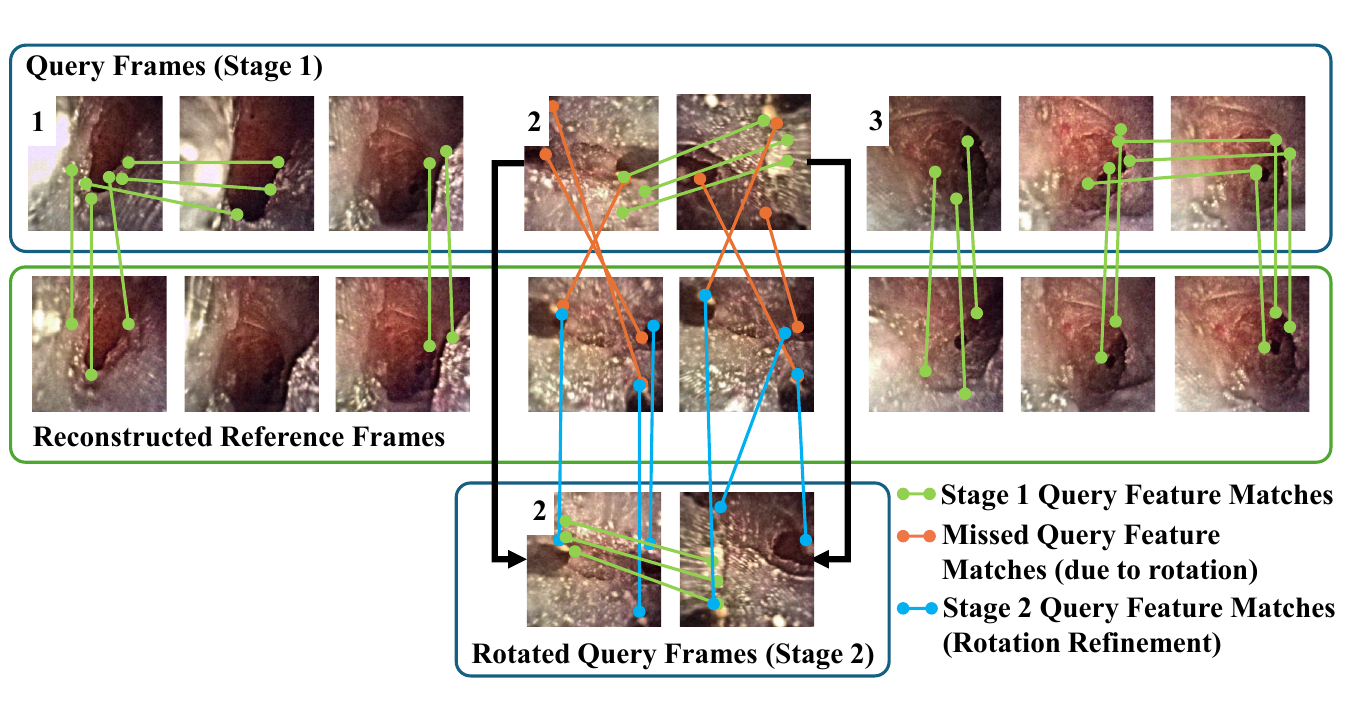}}
\caption{Illustrative feature-matching stages for an example query exploration video. For clarity, only matches involving query frames are shown. Standard SfM often produces fragmented reconstructions from such videos: local video segments can be reconstructed internally, but are not connected into a single reconstruction. Here, these fragments are shown as segments 1–3. In Stage 1, matching against the reference reconstruction connects segments 1 and 3 to the reference map. Group 2 remains unmatched because its frames exhibit an approximately $180^\circ$ in-plane rotation. In Stage 2, these frames are rotated and matched again, recovering their correspondence to the reference map. After both stages, all query frames with sufficient direct or indirect matches to the reference reconstruction can be localized.}
\label{fig:fig2_detailed_pipeline}
\end{figure*}

We observed two dominant causes of these failures, illustrated in Fig.~\ref{fig:sfm_failure_examples}. First, rapid scope translation introduces motion blur and reduces visual overlap between adjacent frames, substantially degrading feature matching. Second, abrupt in-plane scope rotations, including near-$180^\circ$ ureteroscope roll, can make neighboring views appear dissimilar to image-retrieval and feature-matching algorithms. This can leave video segments unlocalized when the user rotates the view to access specific anatomical regions. To address these two failure modes, we designed a pose-estimation workflow with two complementary strategies: 1) reference-assisted reconstruction and 2) roll-aware rotation refinement, as outlined in Fig.~\ref{fig:fig2_detailed_pipeline}.

In stage 1, we construct a reference 3D map of each phantom from an expert-collected slow exploration, reducing the subsequent task to localization alone. In stage 2, we introduce two strategies to improve localization of exploration videos against the reference map.

\subsubsection{Stage 1. Reference-Assisted Reconstruction}
\paragraph{Stage 1.1 Reference reconstruction from slow exploration}
For each phantom, we collect a ureteroscope exploration video from an expert surgeon instructed to be slow and deliberate. We refer to this video as the reference video. We reconstruct a reference model from the reference video using an SfM pipeline based on hloc~\cite{sarlin2019coarse} and COLMAP. 2D image features (ALIKED~\cite{Zhao2023ALIKED}) are extracted from video frames, and NetVLAD~\cite{arandjelovic016netvladcnnarchitectureweakly} is used to retrieve candidate covisible image pairs. Local feature matches are then established using LightGlue~\cite{lindenberger2023lightglue}. The resulting feature matches are passed to COLMAP, which estimates reference ureteroscope poses and triangulates sparse 3D map points. The resultant reference model consists of a sparse 3D point cloud representing the anatomy, along with the reference ureteroscope poses localized within.

The reconstructed SfM model is then registered to the CT-derived kidney collecting-system segmentation using manual initialization followed by iterative closest point (ICP). By registering the reference reconstruction into the CT anatomical coordinate frame, we scale the monocular reconstruction to the anatomy size.

\paragraph{Stage 1.2 Initial matching for query exploration videos}
We refer to each exploration video to be localized as a query video. For each query video, we generate feature matches against the reference video of the same anatomy, used for subsequent localization. We retrieve candidate covisible image pairs both among the frames of the query video and between the query video and reference video frames. Local features are extracted and matched using the same ALIKED-LightGlue pipeline described above. At this stage, the query video frames are not yet localized; instead, this step builds the initial collection of image feature matches used by the final localization procedure.

\subsubsection{Stage 2. Rotation refinement}
After the initial query-video matching pass, some query video frames have insufficient correspondences to the reference video due to large in-plane scope rotations. To increase correspondence coverage, we generate a $180^\circ$ image-plane-rotated copy of each unmatched query frame and repeat retrieval and feature matching against the reference frames. The additional matches obtained from the rotated image representation are added to the collection of feature matches and flagged as rotation-refined matches. The overall step is shown in Fig.~\ref{fig:fig2_detailed_pipeline}.

\subsubsection{Exploration Localization}
The combined collection of matched features, including both initial and rotation-refined matches, is then passed to COLMAP to localize query frames against the fixed reference model. COLMAP estimates the ureteroscope poses of query frames relative to the reference model while preserving the reference map geometry. Each successfully localized query-frame ureteroscope pose is then transformed into the CT anatomical coordinate frame using the precomputed reference-map-to-CT registration. The temporally ordered sequence of localized query-frame ureteroscope poses defines the ureteroscope trajectory.


\subsection{Navigation Metrics for Skill Assessment}
We quantify ureteroscope navigation using navigation metrics derived from the recovered ureteroscope poses. The metrics correspond to technical-skill domains commonly assessed in OSATS, including instrument handling, economy of motion, and task efficiency~\cite{Martin1997OSATS}. Specifically, we compute four metrics: translational and angular log dimensionless jerk (LDLJ) ($LDLJ_\mathrm{t}$, $LDLJ_\mathrm{a}$)~\cite{Singh2021MotionSmoothnessCannulation}, normalized path length, and normalized exploration duration. Translational and angular LDLJ quantify the smoothness of ureteroscope motion. Total trajectory path length is normalized by the collecting-system centerline length to account for differences in anatomical scale and reflects the economy of motion, which is the amount of ureteroscope motion required for an exploration. Total exploration duration is similarly scaled by centerline length and reflects the time required to explore a kidney collecting system of a given scale, hence task efficiency.

\subsubsection{Motion Smoothness}
We compute LDLJ~\cite{Singh2021MotionSmoothnessCannulation} separately for translational and rotational ureteroscope motion ($LDLJ_\mathrm{t}$, $LDLJ_{\mathrm{a}}$) to quantify motion smoothness. LDLJ normalizes jerk with respect to motion duration and trajectory length, making it less sensitive to differences in overall task speed or distance traveled. Lower $LDLJ_\mathrm{t}$ and $LDLJ_\mathrm{a}$ indicate smoother motion.

\begin{equation}
\mathrm{LDLJ_\mathrm{t}}
=
\log\left(
\frac{
T^5}{
D^2
}
    \int_{0}^{T}
\left\lVert
\frac{d^3 \mathbf{p}(t)}{dt^3}
\right\rVert^2
\, dt
\right),
\end{equation}

\begin{equation}
\mathrm{LDLJ}_{\mathrm{a}}
=
\log\left(
\frac{
T^5}{
\Theta^2
} \int_{0}^{T}
\left\lVert
\frac{d^3 \boldsymbol{\theta}(t)}{dt^3}
\right\rVert^2
\, dt
\right),
\end{equation}

where $T$ is the total duration of the exploration, $D$ is the total path length, $\Theta$ is the accumulated geodesic rotation angle along the orientation trajectory, and $\mathbf{p}(t)$ and ${\theta}(t)$ are the ureteroscope position and rotation throughout the trajectory. 

\subsubsection{Economy of Motion}
We present an anatomy-normalized economy of motion metric to evaluate operator skill. Because all participants are instructed to fully explore the same kidney anatomies, the exploration path length serves as a relevant proxy for task efficiency.

To account for the differences in the sizes of the kidney anatomies, we normalize the path lengths by the centerline length of each kidney phantom as a proxy of the standard exploration length required. We first compute the exploration path length from the reconstruction,
\begin{equation}
    L = \sum_{i=2}^{N} \| \mathbf{p}_{i} - \mathbf{p}_{i-1} \|_2 ,
\end{equation}
where $\mathbf{p}_{i}$ is the reconstructed ureteroscope position at frame $i$. Shorter path length indicates less redundant traversal and more efficient navigation through the collecting system.

We follow~\cite{lu2024ava} for automated centerline extraction of kidney phantoms. The CT-derived kidney segmentation is converted to a smoothed surface mesh, after which a wave-propagation skeletonization algorithm~\cite{schreiber2006optimal} contracts equal-distance mesh layers into center points. The connected center points form a branching centerline of the renal pelvis and calyces. We use the Skeletor~\cite{skeletor} package.

Lastly, we compute the unitless normalized path length, 
\begin{equation}
    L_\mathrm{norm} = \frac{L}{L_{\mathrm{centerline}}}.
\end{equation}

\subsubsection{Procedural Efficiency}

We use the total task duration to quantify procedural efficiency, 
\begin{equation}
    T = t_{N} - t_{1},
\end{equation}
where $t_{1}$ and $t_{N}$ denote the timestamps of the first and last frames in the analyzed exploration sequence. Shorter task duration indicates faster completion of the exploration task.

Similar to path length, we also adjust task duration by the centerline path length to account for size variation, resulting in
\begin{equation}
    T_{\mathrm{adj}} = \frac{T}{L_{\mathrm{centerline}}}.
\end{equation}

\section{Experimental Setup}
\subsubsection{Video Data Collection Setup}
We made 10 silicone-based phantoms based on patient CT of anatomically normal kidneys, following~\cite{AcarAyberk2025NNAR}, with BegoStone fake kidney stones (Bego, USA) inserted. We CT-scanned the phantoms and segmented the CT volumes to account for differences caused by manufacturing. 

\subsubsection{Reference Exploration Collection}
One expert surgeon performed a thorough, slow-speed reference exploration of each phantom, which was used for generating the reference reconstructions. 

\subsubsection{Reconstruction Quality Verification} 
We asked the expert to also perform explorations on the ten phantoms, at a clinically realistic speed. During this, we attached an electromagnetic (EM) sensor on the ureteroscope tip to provide ground-truth ureteroscope poses for reconstruction quality evaluation. The sensor was tracked using the Aurora EM tracking system (Northern Digital Inc, Canada) to provide 5-degree-of-freedom ureteroscope pose data. This setup was not repeated for trainees as the additional sensor bulk interfered with scope operation, especially for trainees, which can adversely affect skill assessment.


To evaluate the reconstruction pipeline, we perform three complementary verifications. First, for EM-tracked explorations, we compute the translational and rotational root mean square errors (RMSEs) of reconstructed ureteroscope poses, with axial roll excluded from the rotational error as it is unobservable by the EM sensor. Second, we measure the percentage of successfully localized video frames, since reconstruction methods may fail to register a substantial fraction of frames in challenging endoscopic videos, which can adversely impact the derived navigation metrics. 
Finally, after ICP registration of the SfM reconstruction to the CT-derived kidney segmentation, we compute point-cloud alignment metrics to evaluate reconstruction accuracy.

We also included ablation results for comparison. In cases where a method produced multiple smaller reconstructions for a single video, we picked the single reconstruction with the most localized features. Combining these reconstructions would require manual intervention for each exploration, whereas all methods were evaluated as fully automated pipelines without per-exploration adjustment as the intended workflow.

\subsubsection{Comparison of Metrics Across Skill Levels} 
We collected query exploration videos recorded from 10 surgical trainees. 5 trainees had a case count below 100, with Postgraduate Year (PGY) 1-2, and the other five were of PGY 3-5 with a case count above 100. We refer to them as low PGY and high PGY, respectively. All trainees are residents from the urology department at Vanderbilt University Medical Center (VUMC), USA. 

Each trainee explored the 10 phantoms while guided by an expert surgeon to ensure full anatomy exploration, following clinically realistic procedures. This study was approved under VUMC’s Institutional Review Board (IRB \#231997), and informed consent was obtained from all participants. 

To assess whether the navigation metrics computed from the reconstructed poses capture inter-group differences, we compare each metric between the high and low PGY groups using the Mann-Whitney U test. Statistical comparisons involving the expert are not performed because only one expert was included in the study, and the aforementioned sensor attachment factor can also impact the results.

\section{Results}

\begin{table*}[t]
\centering
\caption{Quantitative comparison of reconstruction and localization pipelines.}
\label{tab:reconstruction_pipeline_comparison}
\renewcommand{\arraystretch}{1.25}
\setlength{\tabcolsep}{6pt}
\resizebox{\textwidth}{!}{%
\begin{tabular}{lccccc}
\toprule
\textbf{Reconstruction Pipeline} &
\textbf{\begin{tabular}[c]{@{}c@{}}Mean Pose\\ Translation RMSE (mm)\end{tabular}} &
\textbf{\begin{tabular}[c]{@{}c@{}}Mean Pose\\ Rotation RMSE ($^\circ$)\end{tabular}} &
\textbf{\begin{tabular}[c]{@{}c@{}}Pose Localization\\ Coverage (\%)\end{tabular}} &
\textbf{\begin{tabular}[c]{@{}c@{}}Reconstruction\\ Mean Point Error (mm)\end{tabular}} &
\textbf{\begin{tabular}[c]{@{}c@{}}Reconstruction 95\%\\ Hausdorff Distance (mm)\end{tabular}} \\
\midrule
SfM Pipeline 
& $0.6 \pm 0.2$ 
& $\mathbf{6.3 \pm 1.8}$
& $50.5 \pm 14.9$ 
& $0.8 \pm 0.2$ 
& $37.8 \pm 21.8$ \\

Reference-Assisted Reconstruction 
& $\mathbf{0.5 \pm 0.1}$  
& $7.2 \pm 1.8$
& $80.2 \pm 10.0$ 
& $0.8 \pm 0.1$ 
& $\mathbf{19.9 \pm 20.0}$ \\

Reference-Assisted Reconstruction + Rotation Refinement (RAUL)
& $\mathbf{0.5 \pm 0.1}$ 
& $7.6 \pm 1.6$
& $\mathbf{86.1 \pm 7.2}$ 
& $0.8 \pm 0.2$ 
& $20.0 \pm 20.0$ \\
\bottomrule
\end{tabular}%
}
\end{table*}
\subsection{SfM Reconstruction and Ureteroscope Pose Estimation Validation}
The results in this section are based on 9 EM-tracked expert query exploration videos. We failed to record 1 EM-tracked expert query exploration video due to setup issues.
Table \ref{tab:reconstruction_pipeline_comparison} presents the quantitative results of reconstruction quality with ablation experiments: 1) baseline SfM without reference reconstruction; 2) reference-assisted reconstruction without rotation adjustment, and 3) the complete pipeline. 




\subsubsection{Ureteroscope Pose Estimation Accuracy}
The mean translation RMSE for our method was 0.5 mm (per-exploration RMSE range: 0.36-0.74 mm). The maximum per-frame position error across all trajectories was 2.2 mm. The translation RMSE is consistently low across the ablation configurations. The mean rotation RMSE increases slightly for RAUL compared to the baseline SfM method (7.6 $^\circ$ vs. 6.3 $^\circ$).

\subsubsection{Percentage of Frames Localized}
The mean percentage of images localized by RAUL is 86.1\%, an increase over the 50.5\% coverage of a baseline SfM method, with reduced standard deviation (7.2\% vs. 14.9\%). RAUL also showed modest improvement over reference-assisted only reconstruction (80.2\%), with a reduction in standard deviation (7.2\% vs. 10.0\%).   

\subsubsection{Reconstruction Point Cloud Model Accuracy}
Lastly, the mean point-wise Euclidean distance after ICP is 0.8 mm for all ablation configurations. The mean 95\% Hausdorff distance is 20 mm for RAUL and 19.9 mm for reference-assisted only reconstruction. The baseline SfM method had a much higher distance of 37.8 mm. 

\begin{figure}
    \centering
    \includegraphics[width=0.9\linewidth]{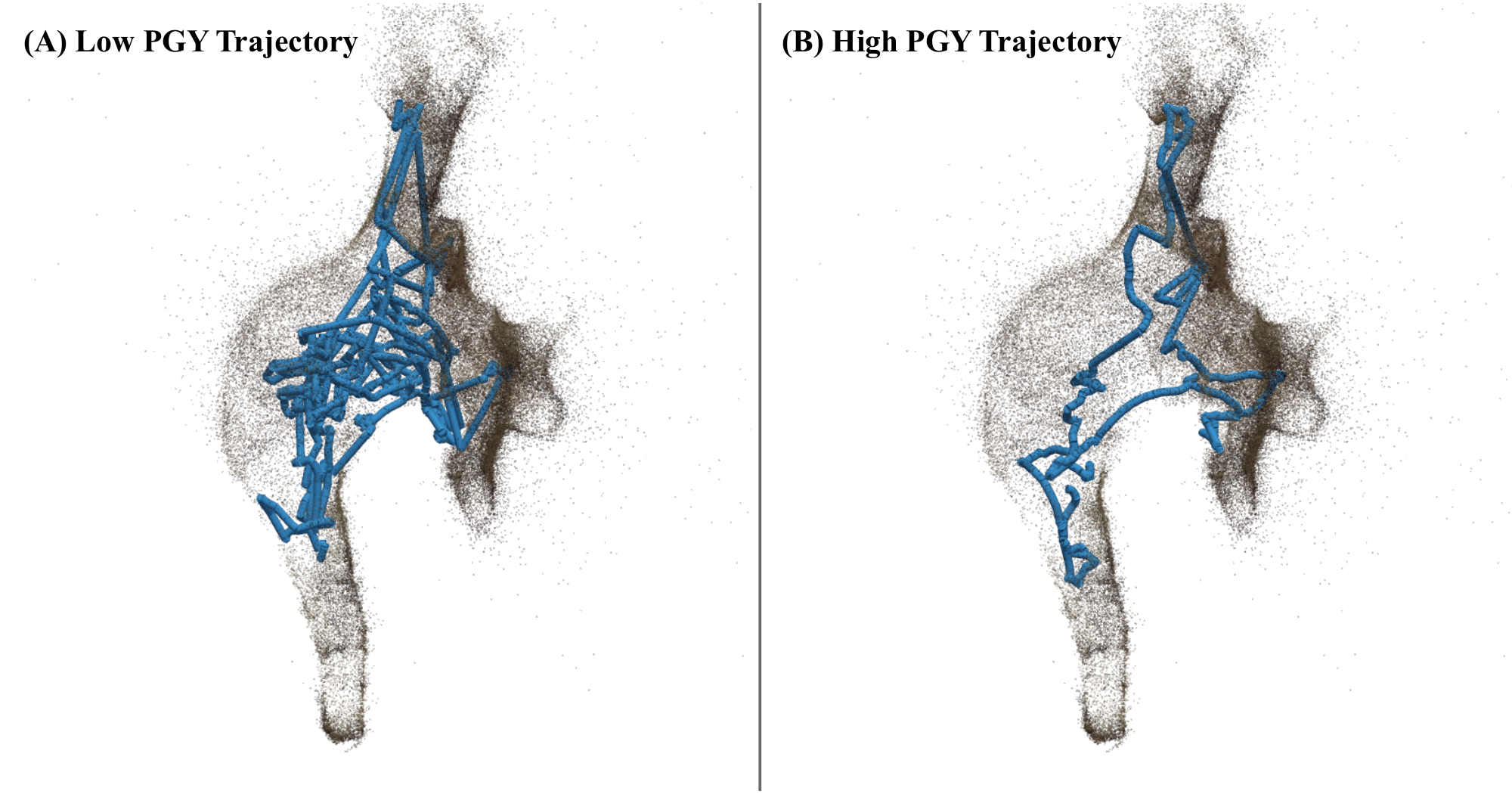}
    \caption{Qualitative comparison of reconstructed trajectories from a low PGY and a high PGY case on the same phantom. The reconstructed low PGY trajectory shows denser local revisits and less direct traversal, whereas the high PGY trajectory shows a more spatially organized and efficient exploration pattern. Reconstruction preserves qualitative differences in navigation behavior across skill levels.}
    \label{fig:high_low_pgy_trajectory}
\end{figure}

\subsection{Skill Assessment Across Skill Levels}
We failed to record 2 high PGY trainee query exploration videos due to setup issues. This left us with 48 high PGY trainee videos and 50 low PGY trainee videos for analysis. Group differences were assessed using Mann-Whitney U tests with rank-biserial effect sizes for pairwise comparisons between high PGY and low PGY groups. The detailed quantitative results are available in Table \ref{tab:motion_summary}.

\subsubsection{Smoothness}
\begin{figure}
    \centering
    \includegraphics[width=\columnwidth]{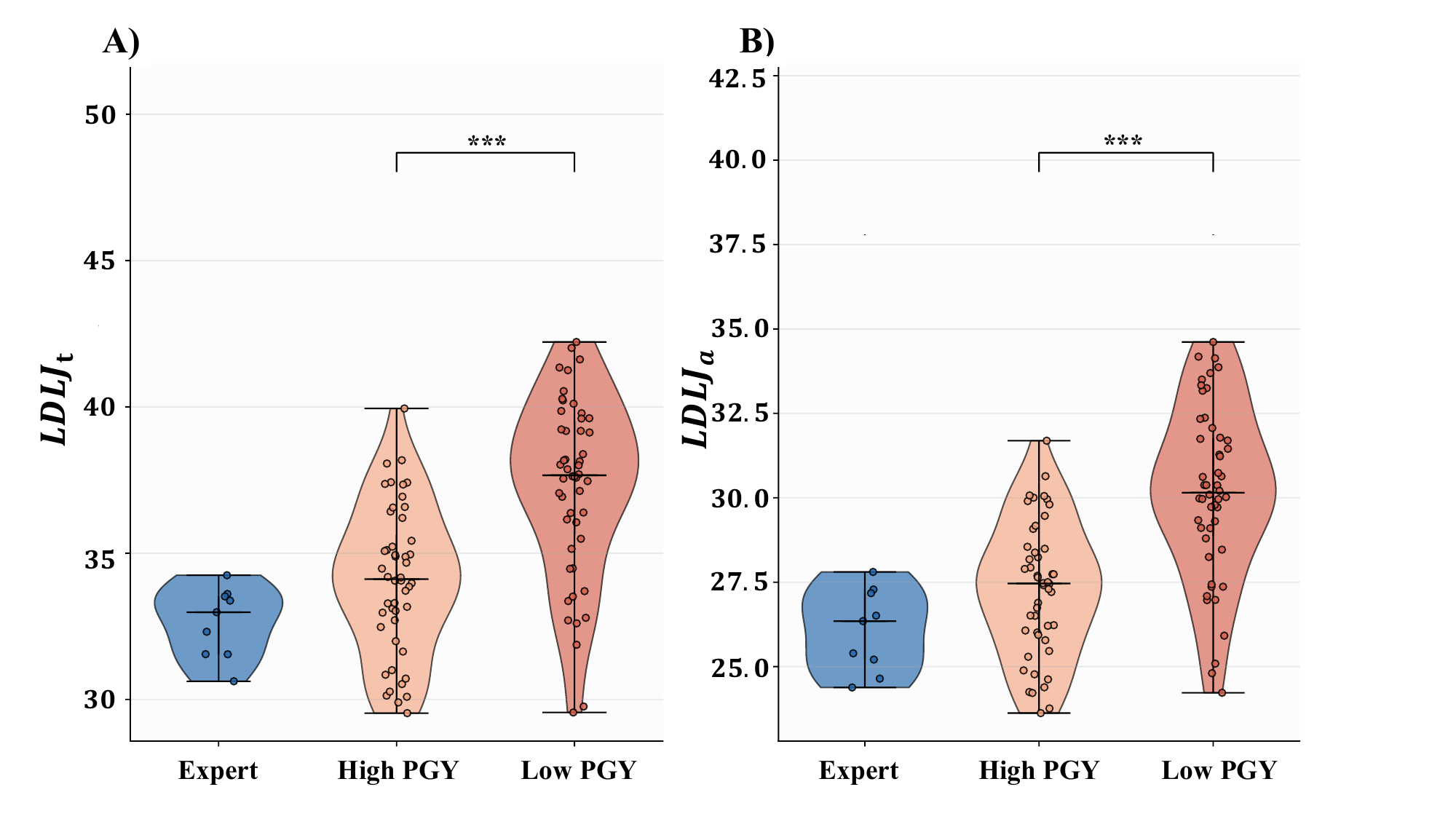}
    \caption{A)  $LDLJ_\mathrm{t}$ across skill groups. B)  $LDLJ_\mathrm{a}$ across skill groups. For both, lower LDLJ values indicate smoother motion. Both translational and angular LDLJ were significantly lower in high PGY cases than in low PGY cases, with angular LDLJ showing the strongest overall group separation. Expert cases also exhibited lower values. ***: $p<0.001$} 
    \label{fig:ldj_results}
\end{figure}

We show the  $LDLJ_\mathrm{t}$ and $LDLJ_\mathrm{a}$ results in Fig.~\ref{fig:ldj_results}. High and low PGY groups are significantly separated by $LDLJ_\mathrm{t}$ ($p<10^{-6}$, $r=+0.58$) and $LDLJ_\mathrm{a}$ ($p<10^{-6}$, $r=+0.62$). This suggests that the high PGY group exhibits much smoother control of the scope motion.

\subsubsection{Economy of Motion} 
\begin{figure}
    \centering
    \includegraphics[width=\columnwidth]{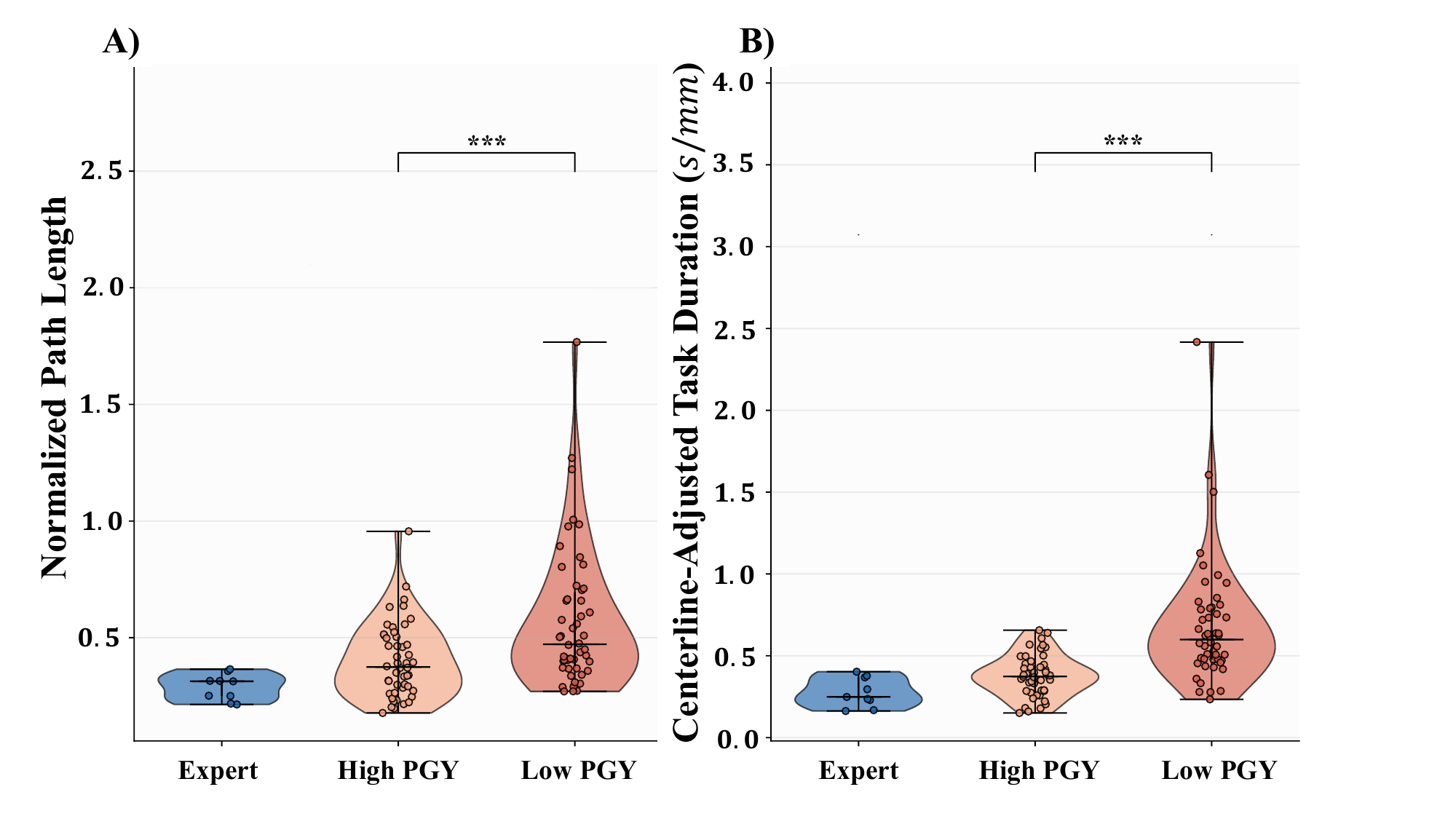}
    \caption{A) Normalized path length across skill groups. B) Centerline adjusted task duration across skill groups. For both, lower values indicate more efficient and faster task completion. Both were lower in Expert and high PGY cases than in low PGY cases. ***: $p<0.001$} 
    \label{fig:path_length_total_duration_results}
\end{figure}

As illustrated in Fig.~\ref{fig:path_length_total_duration_results}, the high PGY group completed the task with significantly shorter normalized path length ($p<10^{-3}$, $r=+0.39$) than the low PGY group, suggesting much more efficient motion. This corroborates the qualitative observations in Fig.~\ref{fig:high_low_pgy_trajectory}. 

\subsubsection{Procedural Efficiency}
Similarly, the high PGY group completed the task with significantly shorter centerline-adjusted task duration ($p<10^{-7}$, $r=+0.67$).

\begin{table}[t]
\centering
\caption{Skill metrics by group. Lower values indicate better performance for all metrics. Values are reported as mean $\pm$ SD.}
\label{tab:motion_summary}
\footnotesize
\renewcommand{\arraystretch}{1.1}
\setlength{\tabcolsep}{2.5pt}
\begin{tabular}{@{}lccc@{}}
\toprule
\textbf{Metric} & \textbf{Expert} & \textbf{High PGY} & \textbf{Low PGY} \\
\midrule
$LDLJ_\mathrm{t}$ 
& $32.9 \pm 1.2$ 
& $34.1 \pm 2.7$ 
& $37.2 \pm 3.0$ \\

$LDLJ_\mathrm{a}$ 
& $25.9 \pm 1.1$ 
& $27.2 \pm 2.1$ 
& $30.1 \pm 2.5$ \\

\shortstack[l]{Normalized Path Length\\$L_\mathrm{norm}$}
& $0.29 \pm 0.06$ 
& $0.40 \pm 0.20$ 
& $0.57 \pm 0.30$ \\

\shortstack[l]{Centerline-Adjusted Duration\\$T_{\mathrm{adj}}$ (s/mm)}
& $0.28 \pm 0.09$ 
& $0.38 \pm 0.12$ 
& $0.68 \pm 0.37$ \\
\bottomrule
\end{tabular}
\end{table}





\section{Discussion}
\subsection{Reference-assisted Reconstruction}
Overall, the proposed method achieved low pose errors comparable to those of the baseline SfM method. Translation errors remained very low at submillimeter level. In contrast, rotation errors were relatively high for both methods. Because this pattern was consistent across the proposed and baseline methods, it may partly reflect uncertainty in the ground-truth sensor attachment rather than reconstruction error alone. Flex or small changes in the relative sensor-to-ureteroscope orientation could produce appreciable angular discrepancies. However, the resultant positional displacement can be small because the sensor was mounted close to the ureteroscope tip, resulting in a very small sensor-to-tip lever arm. This flexing effect may also be exacerbated as more frames are localized, contributing to the slight increase in rotation error for the proposed method. Despite this limitation, navigation metrics derived from both translational and rotational motion consistently distinguished the skill groups, suggesting that the recovered poses were adequate for the present group-level skill assessment.

The primary advantage of the proposed method is improved trajectory localization completeness and robustness rather than improved accuracy on already-localized frames ($86.1\%\ \pm\ 7.2\%$ vs. $50.5\%\ \pm\ 14.9\%$). Trajectory completeness is important for skill assessment. A reconstructed trajectory with low pose error but many missing frames and fragmented segments can still be unsuitable for estimating many navigation behaviors, especially path length and motion smoothness. Missing frames are also unlikely to be random in ureteroscopic video; they often occur during rapid motion (rotation or translation). These difficult segments may contain behaviorally relevant information. Here, the reference-assisted reconstruction ($80.2\%\ \pm\ 10.0$\%) is responsible for the majority of improvement, whereas the roll-aware rotation refinement provided modest gains. By maintaining good pose accuracy while substantially improving frame coverage, the proposed reconstruction strategy provides a more reliable basis for computing downstream navigation metrics.

More broadly, the reconstruction pipeline should be viewed as a technical foundation for objective skill assessment rather than solely as a standalone 3D mapping method. To our knowledge, this study provides the first large-scale demonstration of robust, video-only ureteroscope pose recovery in anatomically realistic kidney phantoms. The evaluation includes more than 100 full-length kidney-exploration videos acquired at realistic operating speeds, each exceeding one minute in duration. These results demonstrate the feasibility of recovering ureteroscope poses without external tracking sensors, providing a practical basis for scalable objective assessment with little additional hardware cost or workflow complexity. Registering the recovered poses to CT-derived anatomy further enables anatomy-aware interpretation of ureteroscope motion, including how efficiently the ureteroscope is navigated through the collecting system. 

The method has a few technical limitations. First, the current pipeline requires close to one hour per query exploration on a modern PC, making it suitable for post-procedure assessment but not real-time feedback. This runtime is largely due to the SfM backend, especially the global bundle adjustment. Future work will investigate alternative reconstruction backends and parameter settings to better balance localization accuracy and runtime. Second, the method is sensitive to differences in ureteroscope image characteristics. Performance can degrade when reference and query videos are acquired using probes with very different image characteristics. This was not a major limitation in our phantom study, where the ureteroscope model was controlled, and may be less problematic in clinical environments with standardized devices; however, improving robustness to probe variability can still be important for broader deployment. 

\subsection{Skill Assessment}
The skill-assessment results indicate that the localized ureteroscope poses capture meaningful differences in ureteroscopic behavior across experience levels. Both smoothness and economy of motion metrics showed significant group-level differences. Expert and high PGY operators generally completed the task with shorter procedure durations and shorter path lengths, suggesting more efficient exploration. Smoothness metrics showed a complementary pattern:  $LDLJ$, particularly  $LDLJ_\mathrm{a}$, consistently separated low and high PGY groups, indicating that less experienced operators exhibited more abrupt and less smooth ureteroscope motion. Together, these findings suggest that the evaluated navigation metrics characterize complementary aspects of ureteroscopic performance. Exploration duration and path length primarily reflect overall navigation efficiency, whereas $LDLJ$ reflects local motion control. 

The qualitative examples in Fig.~\ref{fig:high_low_pgy_trajectory} also show that the recovered trajectories preserve meaningful differences in exploration behavior, including missed versus reached calyces and inefficient, looping paths versus more organized exploration. 

Several limitations remain in the present skill-assessment study. 
First, the dataset includes a modest number of participants and limited variation in experience level. With ten residents in total, analysis beyond binary group comparison was challenging because individual variability could dominate more fine-grained comparisons. Future studies will include more participants across a broader range of experience levels, including additional residents and expert surgeons. Second, this work primarily adopts established navigation metrics to validate the proposed reconstruction pipeline for skill assessment. While effective for initial method validation, these metrics underutilize the anatomy-specific information available from CT-registered endoscope trajectories. Future work will investigate ureteroscopy-specific, anatomy-aware metrics, such as wall-collision avoidance and calyx visitation patterns. Finally, the present analysis focuses on group-level statistical differences; subsequent work will evaluate whether these metrics can support automated skill classification and individualized feedback.

\section{Conclusion}

In this work, we present RAUL, a reference-assisted ureteroscope localization pipeline, used for scalable objective skill assessment. Our proposed pipeline significantly improved localization coverage, while maintaining high localization accuracy. The resulting navigation metrics derived from the recovered trajectories exhibited a statistically significant difference between resident groups with different experience levels. This finding supports the ability of the ureteroscope poses to capture meaningful differences in ureteroscopic behavior. When combined with low-cost, anatomically realistic phantoms, the proposed approach can provide a versatile out-of-OR training and assessment system that enables objective feedback without additional tracking hardware. This supports an affordable and scalable pathway toward quantitative skill assessment in flexible ureteroscopy.

\bibliographystyle{IEEEtran}
\bibliography{references}

\end{document}